\documentclass[lettersize,journal]{IEEEtran}
\usepackage{amsmath,amsfonts}
\usepackage{algorithmic}
\usepackage{algorithm}
\usepackage{array}
\usepackage[caption=false,font=normalsize,labelfont=sf,textfont=sf]{subfig}
\usepackage{textcomp}
\usepackage{stfloats}
\usepackage{url}
\usepackage{verbatim}
\usepackage{graphicx}
\usepackage{amsmath,amssymb} 
\usepackage{cite}
\usepackage[colorlinks,linkcolor=blue]{hyperref}
\usepackage{multirow}

\usepackage{color, colortbl}
\definecolor{mygray}{gray}{0.926}

\begin{document}

\title{Latent Multi-Relation Reasoning for GAN-Prior based Image Super-Resolution}

\author{Jiahui~Zhang, Fangneng~Zhan, Yingchen~Yu, Rongliang~Wu,\\ Xiaoqin~Zhang,~\IEEEmembership{Senior Member,~IEEE}, Shijian~Lu$^*$
\IEEEcompsocitemizethanks{
\IEEEcompsocthanksitem J. Zhang, Y. Yu, R. Wu, and S. Lu are with the School of Computer Science and Engineering, Nanyang Technological University, Singapore.
\IEEEcompsocthanksitem F. Zhan is with the Nanyang Technological University, Singapore and Max Planck Institute for Informatics, Germany.
\IEEEcompsocthanksitem X. Zhang is with the Wenzhou University, China.
}

\thanks{* indicates the corresponding author. Email: shijian.lu@ntu.edu.sg}
}

\maketitle

\begin{abstract}
Recently, single image super-resolution (SR) under large scaling factors has witnessed impressive progress by introducing pre-trained generative adversarial networks (GANs) as priors. However, most GAN-Priors based SR methods are constrained by an attribute disentanglement problem in inverted latent codes which directly leads to mismatches of visual attributes in the generator layers and further degraded reconstruction. In addition, stochastic noises fed to the generator are employed for unconditional detail generation, which tends to produce unfaithful details that compromise the fidelity of the generated SR image. We design LAREN, a \textbf{LA}tent multi-\textbf{R}elation r\textbf{E}aso\textbf{N}ing technique that achieves superb large-factor SR through graph-based multi-relation reasoning in latent space. LAREN consists of two innovative designs. The first is graph-based disentanglement that constructs a superior disentangled latent space via hierarchical multi-relation reasoning. The second is graph-based code generation that produces image-specific codes progressively via recursive relation reasoning which enables prior GANs to generate desirable image details. Extensive experiments show that LAREN achieves superior large-factor image SR and outperforms the state-of-the-art consistently across multiple benchmarks.

\end{abstract}

\begin{IEEEkeywords}
large factor SR, GAN prior, latent multi-relation reasoning, graph-based disentanglement, hierarchical multi-relation reasoning, graph-based code generation.
\end{IEEEkeywords}

\section{Introduction}

\IEEEPARstart{S}{ingle} image super-resolution (SR) aims to reconstruct a realistic high-resolution (HR) image from its low-resolution (LR) counterpart. Thanks to the advance of generative adversarial networks (GANs), image SR has recently achieved remarkable progress by utilizing perceptual and adversarial losses~\cite{sajjadi2017enhancenet}~\cite{wang2018esrgan}~\cite{ledig2017photo}~\cite{johnson2016perceptual}. However, GAN-based SR methods tend to generate various artifacts and unnatural textures while handling large-factor SR such as 8x, 16x, or even 64x SR. This is largely due to the excessive loss of textures and details in the LR input images, which makes it very challenging to generate realistic SR images.

\begin{figure}[t]
\begin{center}
\includegraphics[width=1\linewidth]{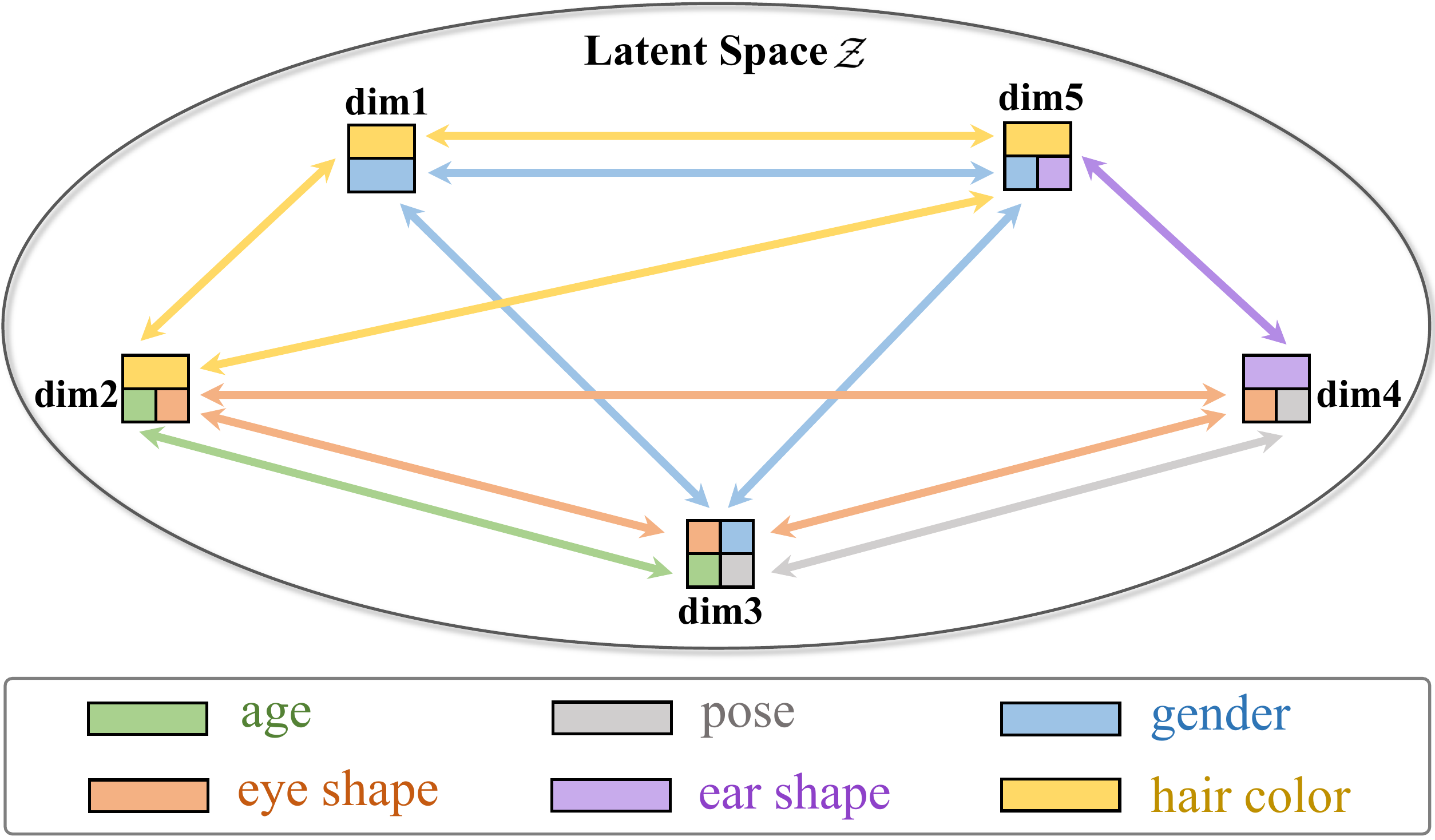}
\end{center}
\caption{
Illustration of multi-level attribute relations across latent dimensions (with five dimensions and six attributes for explanation) in latent space $\mathcal{Z}$: Each latent dimension in the entangled space $\mathcal{Z}$ may capture multiple attributes of different levels (e.g., from global gender and age to hair color). Therefore, there exists multiple relations among these multi-level attributes across dimensions (i.e., multi-level attribute relations). High-quality disentanglement of these attributes in latent space is crucial to the success of GAN-prior based large-factor SR.
}
\label{figure1}
\end{figure}

GAN priors provide rich statistical information from large-scale datasets that facilitates large factor image SR tasks greatly. Due to its powerful generation capability, several studies ~\cite{menon2020pulse}~\cite{gu2020image} adopt a pre-trained StyleGAN ~\cite{karras2019style}~\cite{karras2020analyzing} to provide priors and reconstruct images from the inverted latent codes with large magnification factors. However, these methods often struggle to reconstruct high-fidelity SR images due to the limited expressivity of the latent codes resulting from the low-dimensional input. Recently, \cite{chan2021glean} presents generative latent bank (GLEAN) that introduces multi-scale encoder features for compensating for the limited expressivity of latent codes and generating faithful images toward ground truth at large magnification factors. However, GAN-Prior based SR methods still suffer from two major constraints that prevent them from generating realistic SR images.

\begin{figure*}[t]
\begin{center}
\includegraphics[width=1\textwidth]{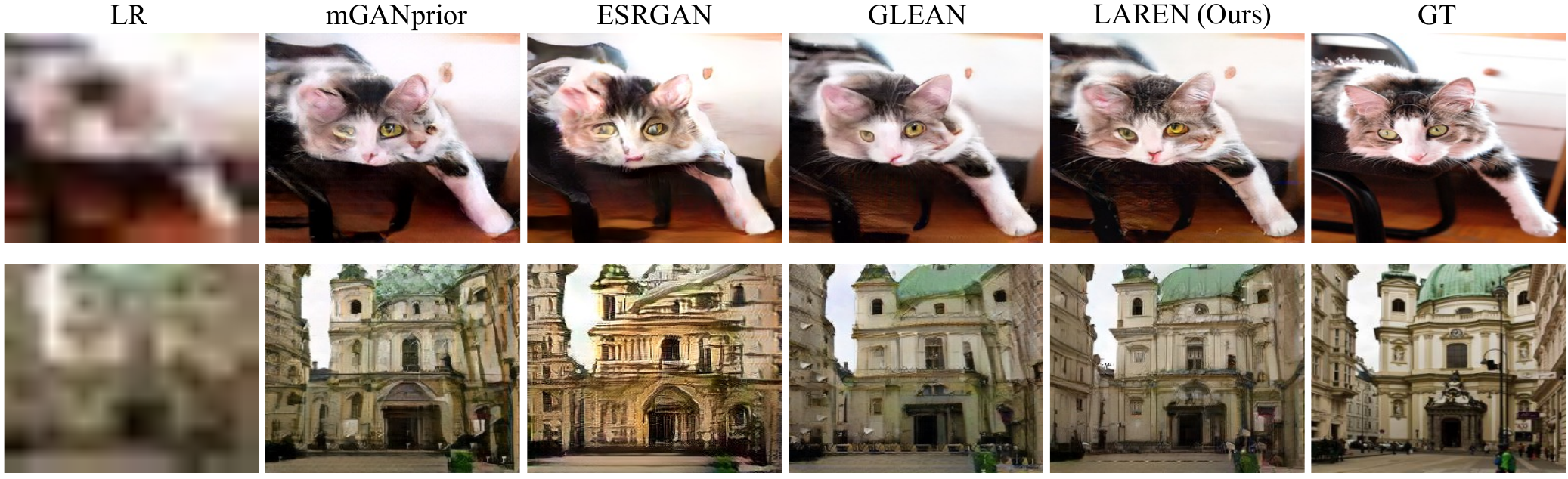}
\end{center}
\caption{\textbf{Illustration of large factor (16 $\times$) super-resolution.} Compared with state-of-the-art mGANprior \cite{gu2020image}, ESRGAN \cite{wang2018esrgan} and GLEAN \cite{chan2021glean}, the proposed LAREN can generate superior SR images with more faithful visual attributes and details.}
\label{teaser}
\end{figure*}

The first constraint lies with insufficient attribute disentanglement of latent codes that directly impairs the generation capability of pre-trained StyleGAN. Specifically, different layers of generator capture image attributes of different levels (i.e., abstract semantics to contents)~\cite{gu2020image}, while the poorly disentangled latent codes may capture mixed attributes of different levels undesirably. Under such circumstance, injecting the entangled latent codes into their corresponding generator layers will lead to visual attribute mismatches, and further compromised reconstruction quality. Existing methods naively employ affine transformation to transform the original entangled latent space $\mathcal{Z}$ to an intermediate latent space $\mathcal{W}$~\cite{karras2019style}, $\mathcal{W+}$~\cite{abdal2019image2stylegan} or $\mathcal{S}$~\cite{wu2021stylespace} which often implicitly produces sub-optimal disentanglement due to the complex multi-level attribute relations across latent dimensions (as illustrated in Fig.~\ref{figure1}) in the entangled space $\mathcal{Z}$. The second constraint lies with stochastic noises that are injected into StyleGAN generator for better texture and detail synthesis in unconditional generation. However, the generation with the stochastic noises is not aligned with single image SR task which is conditional and aims to generate high-fidelity images with respect to ground truth~\cite{chan2021glean}.

We address large-factor image SR challenges by designing \textbf{LA}tent multi-\textbf{R}elation r\textbf{E}aso\textbf{N}ing (\textbf{LAREN}) that exploits graph relation modelling to reconstruct realistic SR images in latent space. LAREN has two innovative designs. The first is a graph disentangled module (GDM) that produces a superior disentangled latent space $\mathcal{G}$ as compared with the existing latent spaces $\mathcal{W}$~\cite{karras2019style}, $\mathcal{W+}$~\cite{abdal2019image2stylegan} and $\mathcal{S}$~\cite{wu2021stylespace}. Intuitively, the entangled latent space $\mathcal{Z}$ can be better disentangled by exploiting multi-level attribute relations across latent dimensions. We thus design hierarchical multi-relation reasoning in GDM that models multi-level attribute relations between any pair of nodes (constructed by the latent code $z$) for better disentanglement. Similar to $\mathcal{W}$~\cite{karras2019style}, $\mathcal{W+}$~\cite{abdal2019image2stylegan} and $\mathcal{S}$~\cite{wu2021stylespace}, the disentangled space G can be learnt in an unsupervised manner. The second is a graph-based code generation module (CGM) that adopts image-specific codes instead of stochastic noises as the input of pre-trained GAN. In particular,  CGM leverages the disentangled hierarchical attributes to obtain image-specific codes for generating faithful details. In addition, we design recursive relation reasoning in CGM to progressively update the image-specific codes in each generator layer, which perfectly aligns with the progressive growing architecture of pre-trained GAN. Thanks to the powerful attribute disentanglement by GDM and faithful code generation by CGM, LAREN achieves superior performance in large factor super-resolution as illustrated in Fig. ~\ref{teaser}.

The contribution of this work is threefold. \textit{First}, we design a graph-based disentangled module that models multi-level attribute relations between each pair of graph nodes via hierarchical multi-relation reasoning, which leads to a better decoupled latent space $\mathcal{G}$ and better attribute disentanglement.  \textit{Second}, we design a graph-based code generation module with a designed recursive relation reasoning to progressively generate image-specific codes instead of stochastic noises, achieving desirable details in single image SR. \textit{Third}, extensive experiments show that the proposed LAREN achieves superior large-factor SR quantitatively and qualitatively over different types of images such as human faces, generic objects, and scenes.

The rest of this paper is organized as follows. Section \ref{related} presents related works. The proposed method is then described in detail in Section \ref{method}. Experimental results are further presented and discussed in Section \ref{experiments}. Finally, concluding remarks are drawn in Section \ref{conclusion}.

\section{Related Work}
\label{related}

\textbf{Image Super-Resolution. } For image SR, one common paradigm trains networks solely with $l_1/l_2$ loss \cite{he2019ode}\cite{zhang2021context}\cite{zhou2020cross}\cite{niu2020single}\cite{dai2019second}\cite{zhang2018image}\cite{dong2014learning}\cite{mei2021image}\cite{magid2021dynamic}\cite{zhang2021blind}. Although this paradigm achieves remarkable PSNR performance, it tends to generate over-smoothing SR images with low visual quality. To resolve this issue, a GAN-based paradigm \cite{zhang2019ranksrgan}\cite{wang2018esrgan}\cite{ledig2017photo}\cite{sajjadi2017enhancenet}\cite{zhan2019spatial}\cite{zhan2021bi}\cite{zhan2021unbalanced}\cite{zhan2022marginal} is proposed for generating realistic images by training the generator with perceptual \cite{johnson2016perceptual}\cite{zhan2022modulated} and adversarial losses. However, this paradigm tends to produce various artifacts and unnatural textures.

Large factor SR ~\cite{zhou2021cross}~\cite{lai2017deep}~\cite{shang2020perceptual}~\cite{dahl2017pixel}~\cite{hyun2020varsr} has attracted increasing attention in recent years. For instance, Hyun et al.~\cite{hyun2020varsr} propose a variational super-resolution network for 8$\times$ SR and restore details by matching the latent distributions of LR and HR images. Zhou et al.~\cite{zhou2021cross} achieve up to 8$\times$ SR by a plane-aware attention mechanism. However, large-factor SR remains a great challenge as the image downsampling (8$\times$, 16$\times$, and even 64$\times$) leads to excessive loss of visual details and texture features in LR images, which significantly aggravates the challenge in reconstructing high-quality SR images.

Recently, GAN inversion becomes a prevalent paradigm in various computer vision tasks ~\cite{pan2021exploiting}~\cite{wang2021towards}~\cite{yang2021gan}\cite{zhan2021multimodal}\cite{yu2022towards}, including large factor SR ~\cite{gu2020image}~\cite{menon2020pulse}~\cite{chan2021glean}. It aims to invert a given image back into the latent space of a pre-trained GAN and then reconstruct the image from the inverted code by a generator. GAN inversion has been successfully incorporated for large-factor image SR, and most existing methods leverage a pre-trained StyleGAN ~\cite{karras2019style}~\cite{karras2020analyzing} to provide strong priors. However, most existing large-factor SR methods (e.g., PULSE~\cite{menon2020pulse}) are constrained by the limited expressivity of the latent codes, which often leads to low fidelity in the generated images. Although GLEAN~\cite{chan2021glean} provides additional multi-resolution features to the pre-trained StyleGAN to mitigate this issue, it still struggles in large-factor SR due to insufficient disentanglement of latent space and the stochasticity of injected noises.

\textbf{Graph Relation Reasoning. }
Graphs have been widely adopted in computer vision community due to its features in non-local operation ~\cite{zhou2020cross}~\cite{mou2021dynamic}~\cite{valsesia2019image}~\cite{wang2018non}~\cite{valsesia2020deep} and relation reasoning ~\cite{santoro2017simple}~\cite{chen2019graph}~\cite{zhu2021semantic}. For example, Santoro et al.~\cite{santoro2017simple} design a simple relational network (RN) to answer relational questions. Chen et al.~\cite{chen2019graph} propose global reasoning unit (GloRe) that exploits via graph convolution to capture global relations among distant regions for image classification. Zhu et al.~\cite{zhu2021semantic} introduce explicit semantic relation reasoning to learn novel object detectors. However, most existing graph development cannot be straightly applied to the latent relation reasoning in GAN-Prior based image SR task largely because they cannot model multiple relations at the level of each individual attribute and also suffer from a lack of relation recursion for progressive generation of image-specific codes.

We explore graph-based relation reasoning in GAN-Prior based SR tasks. Specifically, we design two novel graph-based relational reasoning techniques in latent space. The first is hierarchical multi-relation reasoning that models multi-level attribute relations between any pair of nodes for better disentanglement of the latent space. 
The second is recursive relation reasoning that faciliates the progressive generation of image-specific codes via relation recursion.

\begin{figure*}[t]
\begin{center}
\includegraphics[width=1\textwidth]{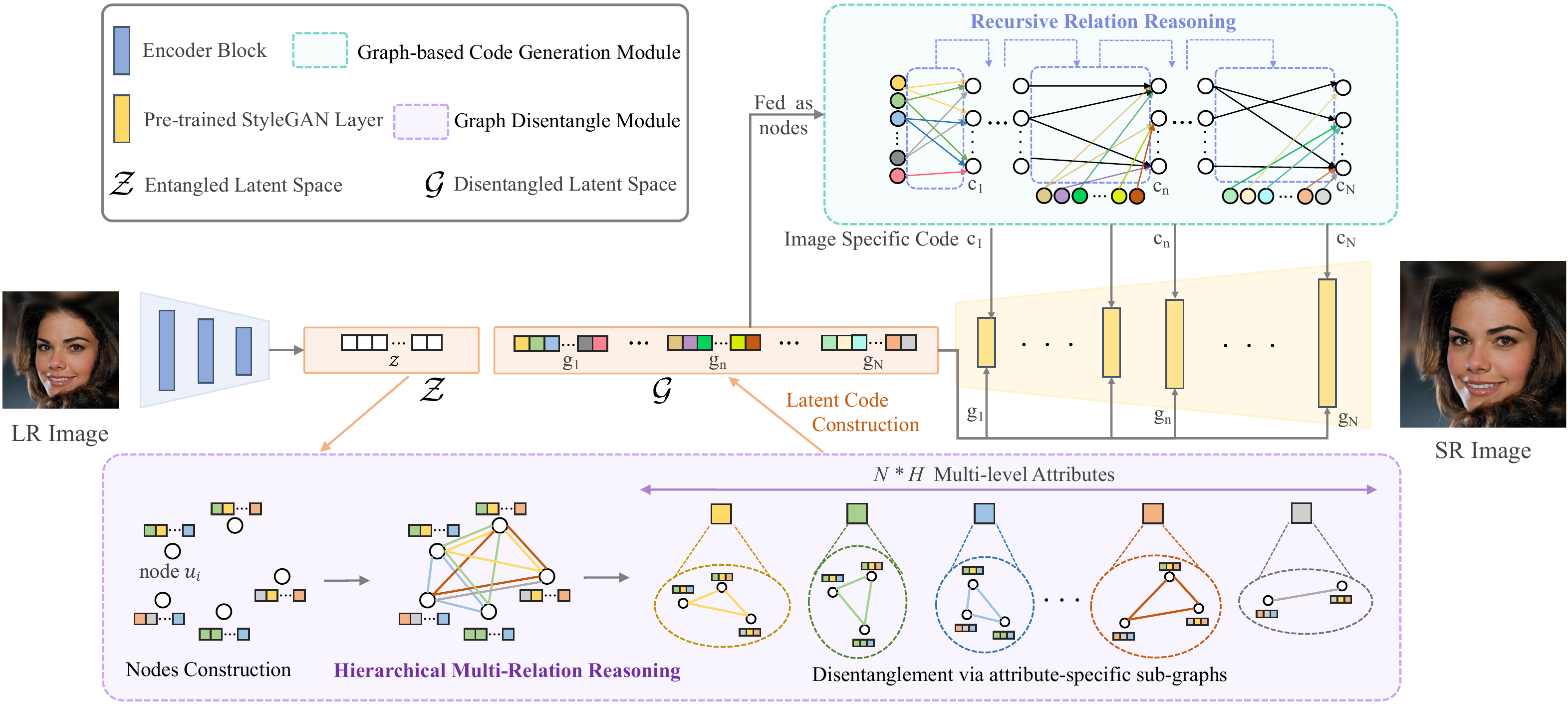}
\end{center}
\caption{
\textbf{The architecture of the proposed LAREN:} LAREN has two innovative designs including a graph disentangled module (GDM) and a graph-based code generation module (CGM). GDM models multi-level attribute relations between any pair of nodes (constructed by the latent code $z$) via hierarchical multi-relation reasoning, and then constructs attribute-specific sub-graphs to capture attributes of different levels which are treated as latent dimensions in the disentangled latent space $\mathcal{G}$ (Each color represents one type of visual attribute). A latent code $g_{n} (n \in [1, N])$ as the input of the $n$-th layer of StyleGAN is constructed by concatenating $H$ latent dimensions. CGM is designed to progressively generate image-specific codes ([$c_{1}$, ... , $c_{n}$, ... , $c_{N}$]). With the proposed recursive relation reasoning, the reasoning at the current layer is conditioned on the relations modelled in the previous layer in CGM.
}
\label{laren}
\end{figure*}

\section{Proposed Method}
\label{method}

\subsection{Overall Framework}

The framework of the proposed LAREN is illustrated in Fig.~\ref{laren}. Following ~\cite{gu2020image}~\cite{chan2021glean}, we introduce GAN inversion and leverage StyleGAN ~\cite{karras2019style}~\cite{karras2020analyzing} to provide strong priors for large-factor SR reconstruction. The input LR image is first encoded to latent code $z \in \mathcal{Z}$, and the latent space $\mathcal{Z}$ is then transformed to a superior disentangled latent space $\mathcal{G}$ by our designed GDM with hierarchical multi-relation reasoning. The well-decoupled latent codes ([$g_{1}$, $g_{2}$, ... , $g_{N}$] $\in \mathcal{G}$) are not only fed to the pre-trained StyleGAN for better exploitation of the GAN priors, but also employed to yield image-specific codes ([$c_{1}$, $c_{2}$, ... , $c_{N}$]) progressively by our designed CGM with recursive relation reasoning. The image-specific codes replace the input stochastic noises which help achieve more desirable details in single image SR with large scaling factors. Details of GDM with hierarchical multi-relation reasoning as well as CGM with recursive relation reasoning will be discussed in the ensuing subsections.

\subsection{Graph Disentangled Module}

Since different layers of pre-trained StyleGAN provide priors of different levels~\cite{gu2020image}, the entangled latent codes where each latent dimension tends to capture multi-level attributes may cause attribute mismatches with the priors of different StyleGAN layers, leading to compromised SR. Although earlier studies ~\cite{karras2019style}~\cite{abdal2019image2stylegan}~\cite{wu2021stylespace} naively construct disentangled spaces by affine transformation, they cannot decouple the latent space effectively due to the complex multi-level attribute relations (illustrated in Fig.~\ref{figure1}). Intuitively, the multi-level attribute relations across latent dimensions will facilitate the attribute disentanglement. We thus propose a graph disentangled module (GDM) to model the multi-level attribute relations to assist transforming $\boldsymbol{z} \in \mathbb{R}^H$ into $N$ latent codes ([$\boldsymbol{g_{1}}$, ... , $\boldsymbol{g_{n}}$, ... , $\boldsymbol{g_{N}}$], $\boldsymbol{g_{n}} \in \mathbb{R}^H$) to match $N$ layers of StyleGAN (affiliated in the disentangled latent space $\mathcal{G}$). 

To derive the multi-level attribute relations, we represent the latent code $\boldsymbol{z}$ as a multi-relational graph $G = (U, E)$, where $U$ , $E$ denote a node set and an edge set within a graph $G$. In this work, we represent the graph $G$ as a set of relational triplets $\{(\boldsymbol{u,r,v})\} \subseteq U\times E\times U$, where $\boldsymbol{r}$ represents the relation embedding between node pairs. The node pair $(\boldsymbol{u} \in \mathbb{R}^C, \boldsymbol{v} \in \mathbb{R}^C) $ is constructed from the latent code $\boldsymbol{z} \in \mathbb{R}^{H}$ by $\boldsymbol{u} = \boldsymbol{W^{u}}  \boldsymbol{z}$ and $\boldsymbol{v} = \boldsymbol{W^{v}}  \boldsymbol{z}$, where $\boldsymbol{W^{u}} \in \mathbb{R}^{C \times H}$ and $\boldsymbol{W^{v}} \in \mathbb{R}^{C \times H}$ are the trainable linear projection matrices. 

\begin{figure}[t]
\begin{center}
\includegraphics[width=1\linewidth]{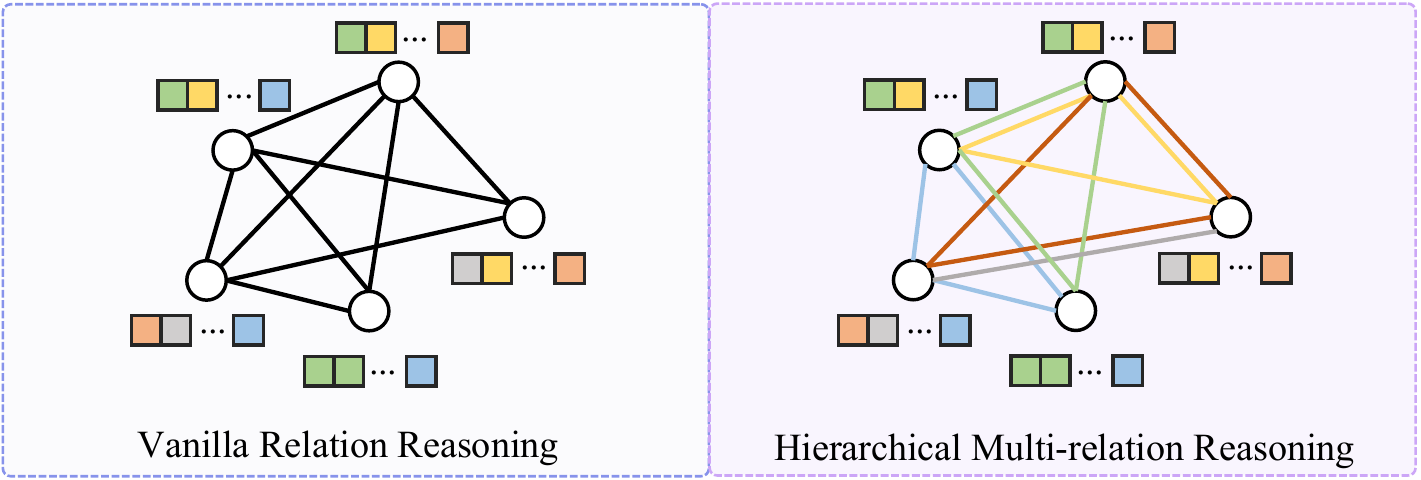}
\end{center}
\caption{
Comparison between \textit{Vanilla Relation Reasoning} (VRR) in ~\cite{zhu2021semantic}~\cite{vaswani2017attention}~\cite{zhang2019self} and our proposed \textit{Hierarchical Multi-Relation Reasoning} (HMRR): unlike VRR that only models the relation between two nodes as a whole, the proposed HMRR models multiple relations at the level of each individual attribute.
}
\label{rr_comparison}
\end{figure}

\textbf{Hierarchical Multi-Relation Reasoning. } The key for graph disentanglement lies with multi-level attribute relation reasoning. Although relation reasoning has been explored in the literature ~\cite{zhu2021semantic}~\cite{chen2019graph} (termed as vanilla relation reasoning), directly applying it can only acquire the relation between a pair of nodes as a whole (as illustrated in Fig.~\ref{rr_comparison}) but cannot model multi-level relations. We therefore propose hierarchical multi-relation reasoning to reason $D$ types of multi-level attribute relations between the node pair separately, where $D = N \times H$ depends on the number of dimensions in the latent space $\mathcal{G}$. The process of simultaneous modelling of multi-level relations can be formulated as:

\begin{equation}
\boldsymbol{r} = \mathrm{Softmax}\bigg(\frac{\delta( \boldsymbol{W_{attr}[u \circleddash v]})}{\lVert \delta(\boldsymbol{W_{attr}[u \circleddash v]}) \rVert_2} + \boldsymbol{P}\bigg),
\end{equation}
where $\boldsymbol{r} \in \mathbb{R}^D$ denotes the relation embedding of the relational triplet, which involves the relations between node pair with respect to $D$ attributes, respectively,  $\circleddash$ represents matrix concatenation operation, $\boldsymbol{W_{attr}} \in \mathbb{R}^{D \times 2C}$ is a trainable matrix for relation modelling with respect to $D$ attributes, $\delta(\cdot)$ represents the activation function ReLU, $\boldsymbol{P} \in \mathbb{R}^{D}$ is embedding for setting zero values to $-\infty$ before softmax function.

With hierarchical multi-relation reasoning, the graph $G$ can be decomposed into $D$ attribute-specific sub-graphs as shown in Fig~\ref{laren}. We adopt the widely used graph embedding method ConvKB~\cite{nguyen2017novel} as attribute extractor $\boldsymbol{f}$ that extracts features from sub-graphs and treat the extracted features as latent dimensions of the disentangled latent space $\mathcal{G}$. The attribute extractor $f$ works on $D$ sub-graphs simultaneously which can be formulated by:

\begin{equation}
\boldsymbol{f} = \delta(Conv([\boldsymbol{u} \circleddash \boldsymbol{r} \circleddash \boldsymbol{v}]))
\end{equation}
Finally, the obtained $D$ latent dimensions are divided into $N$ disentangled latent codes, each of which consists of $H$ latent dimensions. Note similar to the construction of other latent spaces ~\cite{karras2019style}~\cite{abdal2019image2stylegan}~\cite{wu2021stylespace}, the disentanglement process in the proposed space $\mathcal{G}$ is performed in an unsupervised setting.

\subsection{Graph-based Code Generation Module}

Since the stochastic noises to StyleGAN is used for unconditional detail generation, directly introducing vanilla StyleGAN in SR tends to be sub-optimal for generating details toward the ground truth. We thus design a graph-based code generation module (CGM) that generates image-specific codes ([$\boldsymbol{c_{1}}$, ... , $\boldsymbol{c_{n}}$, ... , $\boldsymbol{c_{N}}$], $\boldsymbol{c_n} \in \mathbb{R}^F$) to replace the stochastic noises for faithful detail generation. CGM involves three major designs as illustrated in Fig~\ref{laren}.

First, we propose to yield the image-specific codes for each GAN layer progressively as StyleGAN adopts a GAN architecture with progressive growing. Second, we utilize disentangled latent codes ([$\boldsymbol{g_{1}}$, ... , $\boldsymbol{g_{n}}$, ... , $\boldsymbol{g_{N}}$], $\boldsymbol{g_{n}} \in \mathbb{R}^{H}$) to guide the generation of image-specific codes. Specifically, for each layer, we introduce new nodes each of which is constructed based on a single dimension of the disentangled latent code fed into this layer. As a single dimension is expected to capture one specific attribute, the correlation between attributes and image-specific codes could be established more easily and accurately. Third, to leverage the visual attributes used in previous layers, we propose recursive relation reasoning which combines the relations in the previous layer to model relations within the current layer. The process can be formulated by:  

\begin{equation}
\boldsymbol{T^{n}} = 
\left\{
\begin{aligned}
& \mathrm{norm}((\boldsymbol{W_Q g_n})(\boldsymbol{W_K g_n})^\top), n = 1\\
& \mathrm{norm}(\boldsymbol{W_t^{n-1}T^{n-1}} \circleddash (\boldsymbol{W_Q g_n})(\boldsymbol{W_K g_n})^\top) \\
\end{aligned}
\right.
\label{recursive}
\end{equation}

As defined in ~\cite{vaswani2017attention}, the correlation between the newly introduced nodes of the $n$-th layer and the construction of $\boldsymbol{c_n}$ is modelled by multiplication of the `queries' and `keys', which are transformed by trainable linear projection matrices $\boldsymbol{W_Q} \in \mathbb{R}^{H \times H}$ and $\boldsymbol{W_K} \in \mathbb{R}^{F \times H}$. As Eqn.~\ref{recursive} shows, our relation reasoning technique considers the effect of previous relations by recursively employing relations in the ($n$-$1$)-th layer to model the relations between the output $\boldsymbol{c_{n-1}}$ of the ($n$-$1$)-th layer and the $\boldsymbol{c_{n}}$. The $\boldsymbol{W_t^{n-1}} \in \mathbb{R}^{F \times L}$ is a learnable weight matrix for the ($n$-$1$)-th layer, where $L$ is equal to $H$ when $n$ is set to 2, otherwise, $L=H+F$. In contrast, the vanilla relation reasoning used in ~\cite{zhu2021semantic}~\cite{vaswani2017attention}~\cite{zhang2019self} tends to make the relations between layers independent of each other, which may lead to inaccurate reasoning at the current layer. 

With the fabricated relation matrix $\boldsymbol{T^n}$ of the $n$-th layer, we obtain the image-specific code $\boldsymbol{c_{n}}$ as follows:

\begin{equation}
\boldsymbol{c_{n}} = 
\left\{
\begin{aligned}
& \delta(\boldsymbol{T^n}\boldsymbol{g_n}), n=1 \\
& \delta(\boldsymbol{T^n}[\boldsymbol{c_{n-1}} \circleddash \boldsymbol{g_n}])\\
\end{aligned}
\right.
\end{equation}

Note that the image-specific codes are constructed from the disentangled latent codes. The former aims to replace stochastic noises for detail generation while the latter is used for reconstructing visual attributes. Both of them are essential inputs to the StyleGAN.

\renewcommand\arraystretch{1.6}
\begin{table*}[t]
\renewcommand\tabcolsep{10.6pt}
\centering 
\small
\caption{
Quantitative comparisons on $16\times$SR with human faces, objects (including cat and car) and scenes (including church, bedroom, tower): The proposed LAREN outperforms the state-of-the-art consistently in both PSNR and LPIPS under different categories. All methods are trained with the same training dataset and batch size in experiments. Note for comparison fairness, we skip the comparison with GPEN~\cite{yang2021gan} and GFP-GAN~\cite{wang2021towards} which are specially designed for human faces instead of general objects and scenes.
}
\label{tab_compare}
\begin{tabular}{l||cc||cc||cc||cc} 
\hline
& 
\multicolumn{2}{c||}{\textbf{mGANprior~\cite{gu2020image}}} & 
\multicolumn{2}{c||}{\textbf{ESRGAN~\cite{wang2018esrgan}}} &
\multicolumn{2}{c||}{\textbf{GLEAN~\cite{chan2021glean}}} &
\multicolumn{2}{c}{\textbf{LAREN (Ours)}}
\\
\cline{2-9}
\multirow{-2}{*}{\textbf{Categories}} 
& PSNR $\uparrow$ & LPIPS $\downarrow$ & PSNR $\uparrow$ & LPIPS $\downarrow$ & PSNR $\uparrow$ & LPIPS $\downarrow$ & PSNR $\uparrow$ & LPIPS $\downarrow$ 
\\\hline

\textbf{Face 16$\times$} & 23.50 & 0.4787 & 26.04 & 0.2993 & 26.71 & 0.2735 & \textbf{26.95} & \textbf{0.2631}                  \\

\textbf{Face 32$\times$} & 19.45 & 0.5192 & N/A & N/A &  22.83 & 0.3397 & \textbf{23.11} & \textbf{0.3288}                 \\

\textbf{Face 64$\times$} & 17.32 & 0.5633 & N/A & N/A & 20.96 & 0.3958 & \textbf{21.28} & \textbf{0.3837}                 \\

\hline

\textbf{Cat  16$\times$} & 16.93 & 0.5621 & 18.89 & 0.3692 & 20.03 & 0.3426 & \textbf{20.15} & \textbf{0.3379}                 \\

\textbf{Church 16$\times$} & 14.78 & 0.5569 & 16.56 & 0.3671 & 17.32 & 0.3485 & \textbf{17.58} &  \textbf{0.3283}                 \\

\textbf{Car 16$\times$} & 14.40 & 0.7388 & 19.01 & 0.3180 & 19.61 & 0.2853 & \textbf{19.88} &  \textbf{0.2729}                \\

\textbf{Bedroom 16$\times$} & 16.19 & 0.5597 & 19.18 & 0.3507 & 19.32 & 0.3415 & \textbf{19.45} & \textbf{0.3304}                 \\

\textbf{Tower 16$\times$} & 15.83 & 0.4982 & 17.61 & 0.3240 & 18.33 & 0.2973 & \textbf{18.49} & \textbf{0.2883}                 \\

  \hline
\end{tabular}

\end{table*}

\begin{figure*}[b]
\begin{center}
\includegraphics[width=1\textwidth]{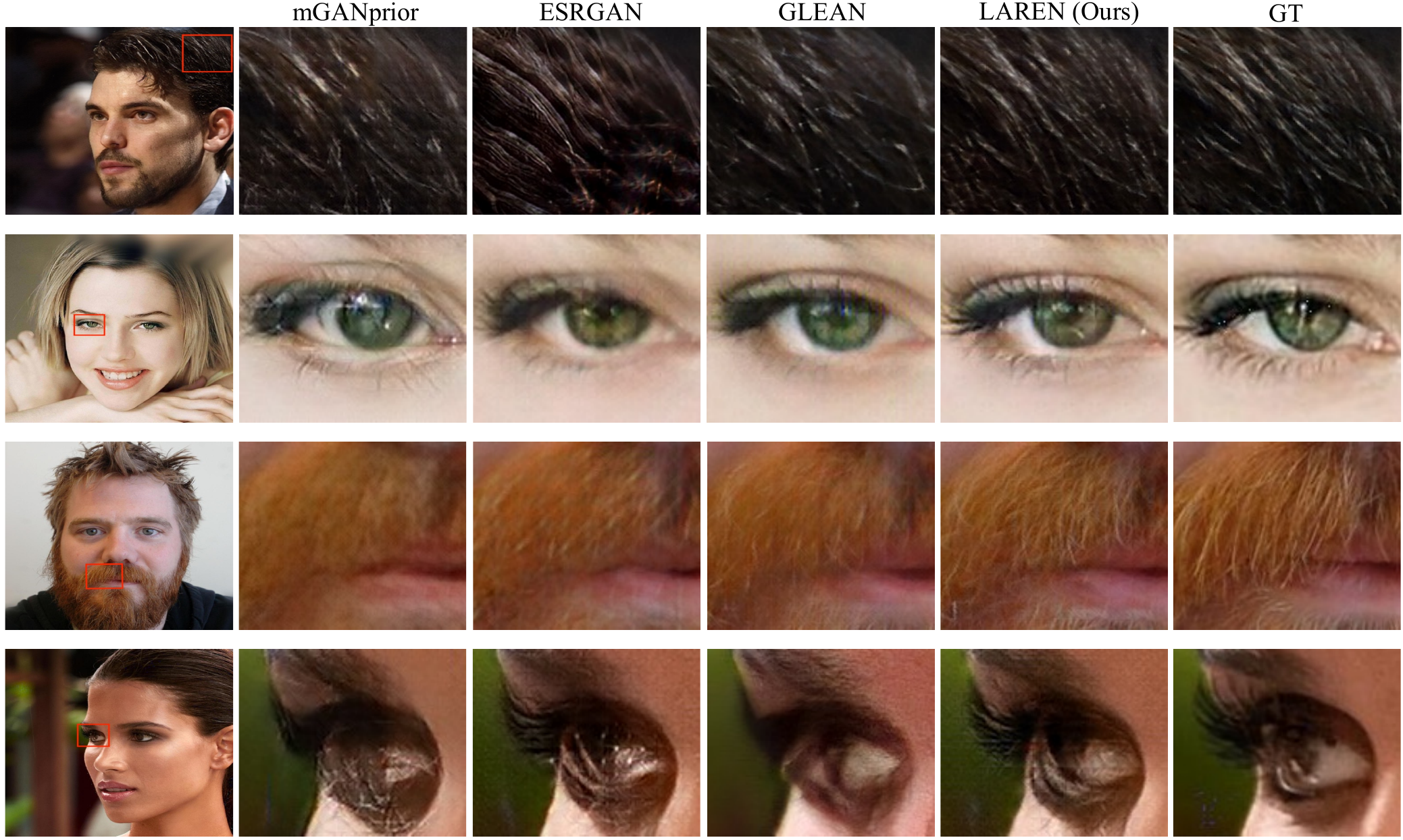}
\end{center}
\caption{Qualitative comparisons of 16$\times$SR on the human face: ESRGAN ~\cite{wang2018esrgan} produces images with unnatural textures. Other GAN-Prior based methods mGANprior ~\cite{gu2020image} and GLEAN ~\cite{chan2021glean} fail to synthesize fine details and faithful attributes due to the insufficient disentangled latent space and stochastic noises provided for pre-trained StyleGAN. The proposed LAREN produces much more realistic and faithful visual attributes and details with our designed GDM and CGM. \textbf{Zoom in for best view.}  }
\label{compare_face}
\end{figure*}

\subsection{Network Details and Loss Functions}

In GDM, the dimension $H$ of the latent code $z$ is 512 which is the same StyleGAN~\cite{karras2019style}. The dimension of the disentangled latent codes ([$g_{1}$, ... , $g_{n}$, ... , $g_{N}$]) is set to 512 as these codes are fed into the StyleGAN, where the value of $N$ is equal to the number of GAN layers. The number of node affiliated in node set $U$ and its dimension are set to $J$=$64$ and $C$=$8$, respectively. In CGM, the dimension $F$ of generated image-specific codes is set to 64. Before feeding it to the pre-trained StyleGAN, we transform each code via linear projection to align its shape with the corresponding StyleGAN layer. The investigation of these parameters is presented in Section~\ref{Para_investi}.

For the loss function, similar to the construction of other disentangled spaces such as $\mathcal{W}$~\cite{karras2019style}, $\mathcal{W+}$~\cite{abdal2019image2stylegan} and $\mathcal{S}$~\cite{wu2021stylespace}, we do not introduce additional losses to constrain the disentanglement process (i.e., in an unsupervised setting) since Karras et al. ~\cite{karras2019style} show that the generator play certain roles in driving disentanglement. The loss function adopted in our model training includes the standard l2 loss, perceptual loss~\cite{johnson2016perceptual}, and adversarial loss, which are the same as existing methods such as GLEAN. The overall model objective $\mathcal{L}$ is a combination of the above losses as follows:
\begin{equation}
\begin{aligned}
\mathcal{L} & = \mathcal{L}_{MSE} + \alpha \mathcal{L}_{per} + \beta \mathcal{L}_{adv}\\
& = \lVert \hat{y} - y\rVert_2^2 + \alpha \lVert \phi(\hat{y}) - \phi(y)\rVert_2^2 + \beta log(1-D(\hat{y})),
\end{aligned}
\end{equation}
where $\hat{y}$, $y$, $\phi(\cdot)$ and $D(\cdot)$ denote the output image, the ground truth, the pre-trained VGG16 network, and the discriminator of StyleGAN, respectively. Meanwhile, $\alpha=0.01$ and $\beta=0.01$ weight the contributions of the perceptual loss $\mathcal{L}_{per}$ and the adversarial loss $\mathcal{L}_{adv}$, respectively.

\begin{figure*}[t]
\begin{center}
\includegraphics[width=1\textwidth]{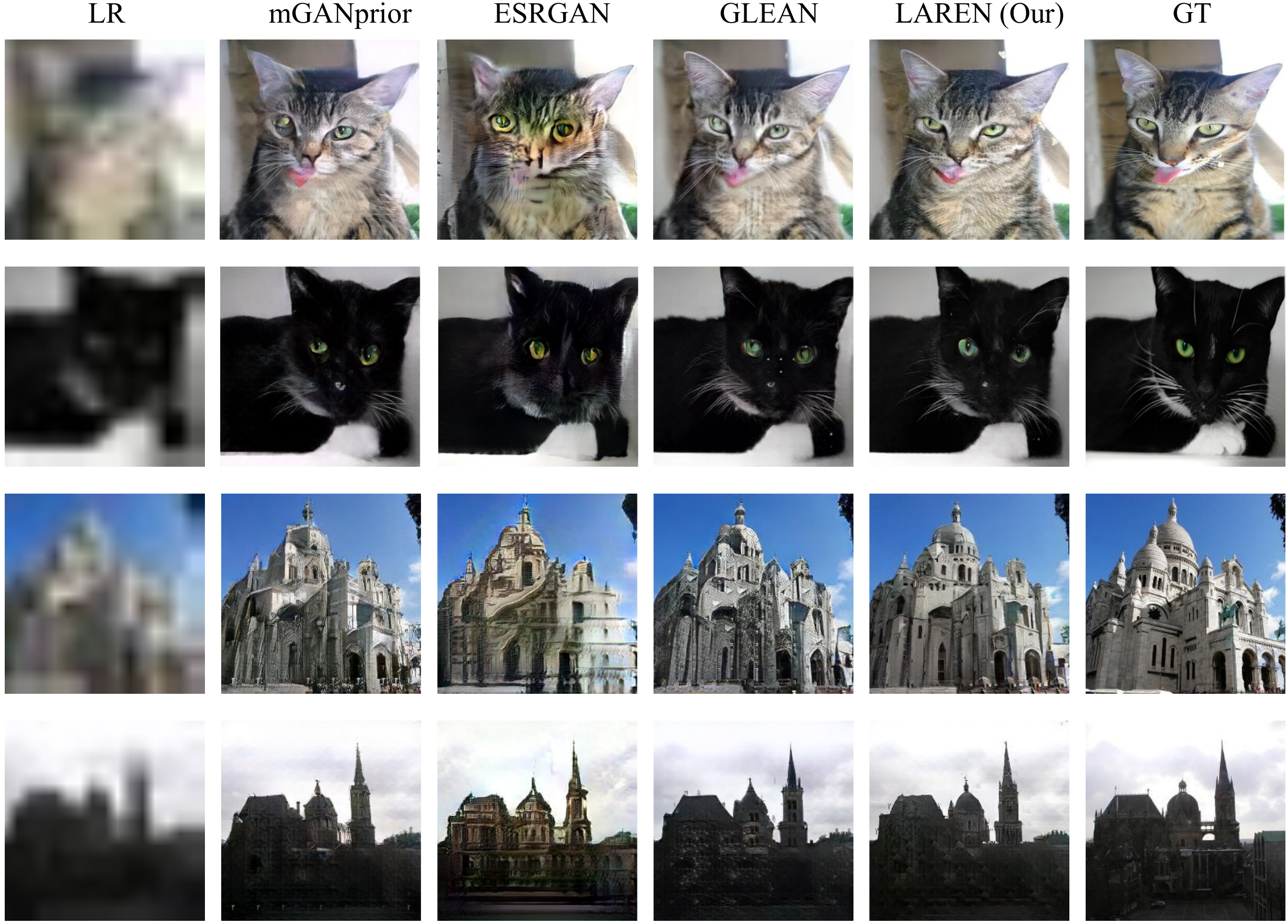}
\end{center}
\caption{Qualitative comparisons of 16$\times$SR on the cat (object) and church (scene): The proposed LAREN outperforms state-of-the-art methods in generating high-fidelity attributes and desirable details. \textbf{Zoom in for best view.}}
\label{compare_other}
\end{figure*}

\section{Experiments}
\label{experiments}

\subsection{Datasets and Implementation Details}

\textbf{Datasets. }
\label{dataset} Following GLEAN ~\cite{chan2021glean}, we conduct experiments on corresponding datasets of various categories, including human faces, cat, church, car, bedroom and tower. 

\noindent
$\bullet$ For human faces, We use FFHQ ~\cite{karras2019style} which consists of $70000$ high-quality face images of resolution $1024^2$ for training. The evaluation is performed on 3000 face images from the testing partition of CelebA-HQ ~\cite{liu2015deep}.

\noindent
$\bullet$ For cat and car, we perform training on the images in the corresponding category of LSUN ~\cite{yu2015lsun} and conduct evaluation on the testing partition of CAT~\cite{zhang2008cat} and CAR ~\cite{krause20133d} respectively since LSUN has no testsets of cat and car.

\noindent
$\bullet$ For outdoor church, bedroom and tower, we conduct training and testing on the corresponding datasets of LSUN.

\textbf{Implementation. }
Similar to GLEAN, we leverage StyleGANv1 ~\cite{karras2019style} or StyleGANv2 ~\cite{karras2020analyzing} (depending on its availability) to provide priors and keep the weights of the pre-trained StyleGAN fixed throughout training without fine-tune. We adopt Adam optimizer with {$\beta_{1}$} = 0.9, {$\beta_{2}$} = 0.999, and $\epsilon$ = {$10^{-8}$} and the initial learning rate is set to 0.0001. The training mini-batch size is set to 8 for human faces and 16 for other categories. We use the Pytorch framework to implement our model with two NVIDIA Tesla V100 GPUs.

\subsection{Comparisons with the State-of-the-Art}

We compare LAREN with three state-of-the-art SR methods over different SR tasks on human faces, objects (cat and car) and scenes (church, bedroom and tower). We set the scale factors from 16 $\times$ to 64 $\times$ for human faces. For other categories, the scale factor is set at $16\times$ only as the resolution of their HR images is very low (e.g., 256$\times$256). Due to the absence of pretrained models, we train mGANprior ~\cite{gu2020image}, ESRGAN ~\cite{wang2018esrgan} and GLEAN ~\cite{chan2021glean} based on their official codes and all methods (including our LAREN) are trained with the same training dataset and batch size in experiments. The tests are over the same whole testset as described in section~\ref{dataset} while computing average PSNR and LPIPS for each category. Table~\ref{tab_compare} shows experimental results. We can see that the proposed LAREN outperforms the three compared state-of-the-art methods consistently in both PSNR and LPIPS across all categories. The superior performance is largely attributed to our designed GDM and CGM that construct a superior disentangled latent space and generate image-specific codes for the pre-trained StyleGAN which enable better use of priors in pre-trained StyleGAN and improve the reconstructed quality. The quantitative experimental results are well aligned with the qualitative results in Figs.~\ref{compare_face} and~\ref{compare_other} where LAREN produces superior SR images with faithful visual attributes and desirable image details. Note for comparison fairness, we did not compare with GPEN ~\cite{yang2021gan} and GFP-GAN ~\cite{wang2021towards} which are specially designed for human faces rather than general objects and scenes.

\begin{table}[!htbp]
  \renewcommand\tabcolsep{10.6pt}
  \centering
  \small
  \begin{center}
  \caption{Verification of our proposed disentangled latent space $\mathcal{G}$  that is constructed by our proposed graph disentangled module: We perform evaluation by using DCI metric~\cite{eastwood2018framework} which consists of three components including `disentanglement' that measures if each latent dimension controls single attribute, `completeness' that measures if each attribute is only captured by one latent dimension, and `informativeness' that measures the classification accuracy of the attributes. The larger the three component values, the better, and the maximum value is 1. }
  \label{disentanglement}
      \begin{tabular}{l|c|c|c|c}
        Latent Space & $\mathcal{W}$ & $\mathcal{W+}$ & $\mathcal{S}$ & $\mathcal{G}$(Ours) \\
        \hline
        \hline
        Disentanglement & 0.47 & 0.56 & 0.68 & \textbf{0.82} \\
        Completeness & 0.51 & 0.63 & 0.76 & \textbf{0.88} \\
        Informativeness & 0.94 & 0.96 & \textbf{0.99} & \textbf{0.99} \\
      \end{tabular}
  \end{center}
\end{table}

\subsection{Discussions}

\textbf{Disentanglement Verification. } We conduct experiments to study the effectiveness of the proposed GDM in decoupling latent space. We adopt DCI (disentanglement/completeness/informativeness) ~\cite{eastwood2018framework} metric to measure the degree of disentangled representations in different latent spaces. Following ~\cite{wu2021stylespace}, we first sample 500K random intermediate latent codes $w \in \mathcal{W}$, transform them to corresponding latent codes $w+ \in \mathcal{W+}$, $s \in \mathcal{S}$ and our $g \in \mathcal{G}$, and then employ StyleGANv2 pre-trained on FFHQ dataset to generate corresponding images, and finally employ classifiers pre-trained on CelebA attributes to annotate images for the training of DCI regressors. As shown in Table~\ref{disentanglement}, the disentanglement and completeness scores of our $\mathcal{G}$ are clearly higher than that of the other two latent spaces, indicating that our proposed GDM has superior capability to construct a decoupled latent space.

\begin{figure*}[!htbp]
\begin{center}
\includegraphics[width=1\linewidth]{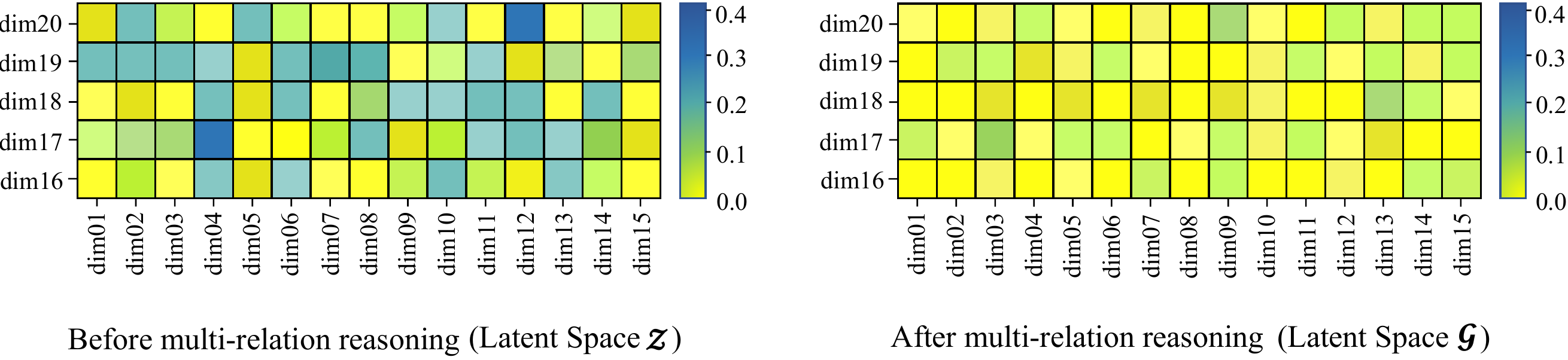}
\end{center}
\caption{
Correlations of original latent dimensions in $\mathcal{Z}$ and disentangled latent dimensions in $\mathcal{G}$ on the CelebA dataset: Due to space limit, we show correlations between a subset of dimensions only ( i.e., between dim 16-20 and dim 1-15,) in both spaces. It shows that the proposed multi-relation reasoning aggregates categorical attribute information in the corresponding latent dimensions effectively and the latent dimensions become less correlated and better disentangled with each other.
}
\label{visualization}
\end{figure*}

\textbf{Visualization of Multi-Relation Reasoning. } Following ~\cite{zhu2021semantic}, we measure the correlations of original latent dimensions in $\mathcal{Z}$ and disentangled latent dimensions in $\mathcal{G}$ as illustrated in Fig.~\ref{visualization}. It can be observed that the proposed multi-relation reasoning suppresses the correlations between latent dimensions in $\mathcal{G}$ significantly. The multi-relation reasoning works as it facilitates attributes propagation across latent dimensions and aggregates attribute information of same category to specific latent dimensions effectively.

\begin{figure*}[!htbp]
\begin{center}
\includegraphics[width=1\textwidth]{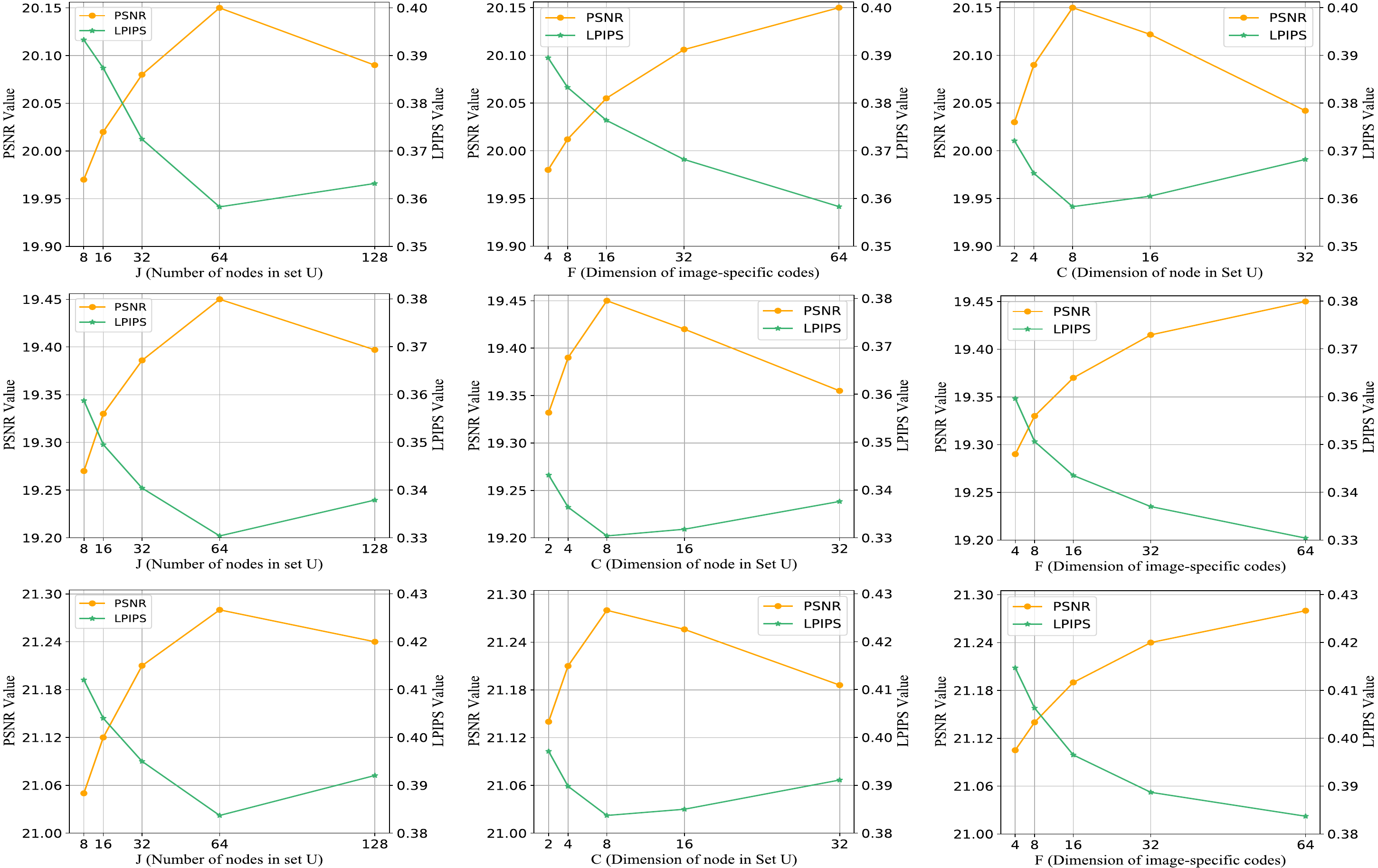}
\end{center}
\caption{
Parameters $J$, $C$, and $F$ in LAREN: The performance of LAREN varies with $J$ (number of nodes in set U), $C$ (dimension of nodes in set U), and $F$ (dimension of image-specific codes). Increasing $F$ improves SR consistently while the optimal $J$ and $C$ lie around 64 and 8. The results displayed from top to bottom are evaluated on 16$\times$ CAT, 16$\times$ bedroom and 64$\times$ face, respectively.
}
\label{para}
\end{figure*}

\textbf{Parameters Investigation}
\label{Para_investi}
We conduct experiments on CAT ~\cite{zhang2008cat}, bedroom and face datasets to study the effect of hyper-parameters in LAREN. For GDM, we mainly study the number of node ($J$) in set $U$ and the node dimension $C$. It can be observed in Fig.~\ref{para} that with the increase of $J$ and $C$, the performance (PSNR $\uparrow$ and LPIPS $\downarrow$ scores) of LAREN is improved. However, the models with excessive nodes and dimensions tend to produce degraded SR, largely due to the increased complexity in multi-relation reasoning. For CGM, Fig.~\ref{para} shows that the performance of LAREN presents positive correlation with the increase of the dimension $F$ of generated image-specific code. However, considering computation costs, we set the dimension $F$ at 64 in LAREN.

\textbf{Model Size Analysis}
We examine the computational complexity and model sizes of the proposed GDM and CGM. As Table \ref{parameter} shows, including GDM increases the model size slightly but improves the model performance significantly. On top of GDM, further including CGM introduces clearly more parameters as well as further performance improvement by a large margin. Overall, the proposed GDM and CGM introduce limited parameters but improve the model performance significantly.

\begin{table}[t]
  \renewcommand\tabcolsep{7.6pt}
  \centering
  \begin{center}
  \caption{The parameter of our baseline model, GDM and CGM. The evaluation is performed over 64$\times$ SR on the CelebA-HQ dataset. It can be seen that our GDM and CGM only introduce a small number of additional parameters while improving the model performance significantly. }
  \label{parameter}
      \begin{tabular}{l|c|c|c}
        \textbf{Model} & Baseline & Baseline+GDM & Baseline+GDM+CGM \\
        \hline
        \hline
        \textbf{Parameter} & 67.1M & 67.1M+0.42M & 67.1M+0.42M+11M \\
        \textbf{PSNR} $\uparrow$ & 20.69 & 21.06 & 21.28  \\
        \textbf{LPIPS} $\downarrow$ & 0.6756 & 0.4668 & 0.3837 \\
      \end{tabular}
  \end{center}
\end{table}

\renewcommand\arraystretch{1.5}
\begin{table}[t]
\small 
\renewcommand\tabcolsep{6pt}
\centering 
\caption{
Ablation studies of LAREN over 64$\times$ SR: The pre-trained StyleGAN in the \textit{Base} model receives latent codes from the latent space $\mathcal{W}$ and stochastic noises. Replacing $\mathcal{W}$ with latent spaces $\mathcal{W+}$ and $\mathcal{S}$ produces \textit{Base+$\mathcal{W+}$} and \textit{Base+$\mathcal{S}$} which outperform the \textit{Base} model consistently. \textit{GDM (HMRR)} and \textit{GDM (VRR)} refer to the GDM with the proposed hierarchical multi-relation reasoning (HMRR) and a GDM variant with vanilla relation reasoning (VRR). \textit{CGM (ReRR)} and \textit{CGM (VRR)} denote CGM with our proposed recursive relation reasoning (ReRR) and a CGM variant with vanilla relation reasoning. The model with both GDM (HMRR) and CGM (ReRR) is equivalent to the proposed LAREN. The evaluations are performed on CelebA-HQ.
}
\label{ablation}
\begin{tabular}{l||cc} 
\hline
& 
\multicolumn{2}{c}{\textbf{Evaluation Metrics}}
\\
\cline{2-3}
\multirow{-2}{*}{\textbf{Models}} 
& PSNR $\uparrow$ & LPIPS $\downarrow$ 
\\\hline

\textbf{Base} & 20.69 & 0.6756         \\

\textbf{Base+$\mathcal{W+}$ } & 20.78 & 0.6369    \\

\textbf{Base+$\mathcal{S}$} & 20.85 &  0.6052            \\

\textbf{Base+GDM (VRR)} & 20.80 &   0.6311           \\

\rowcolor{mygray} 
\textbf{Base+GDM (HMRR)} & 21.06 &    0.4668             \\

\hline

\textbf{Base+GDM (HMRR)+CGM (VRR)} & 21.11 &   0.4572           \\
\rowcolor{mygray} 
\textbf{Base+GDM (HMRR)+CGM (ReRR)} & \textbf{21.28} & \textbf{0.3837}   \\
  \hline
\end{tabular}

\end{table}

\subsection{Ablation Studies on Graph Disentangled Module}

\begin{figure}[t]
\begin{center}
\includegraphics[width=1\linewidth]{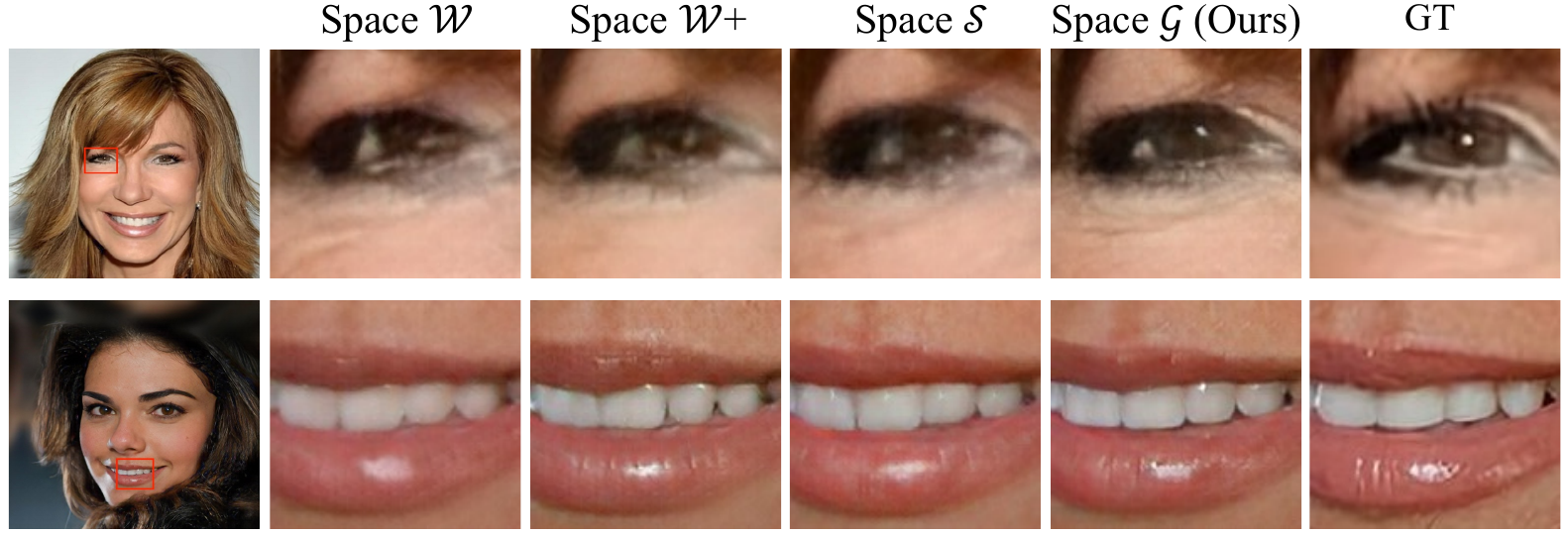}
\end{center}
\caption{
Qualitative ablation studies of the proposed latent space $\mathcal{G}$ on 16 $\times$ SR. The proposed latent space $\mathcal{G}$ offers superior attribute disentanglement, and relevant model can better utilize the priors in pre-trained StyleGAN and generate more faithful visual attributes as compared with latent spaces $\mathcal{W}$, $\mathcal{W+}$ and $\mathcal{S}$ (e.g., eyes and teeth). \textbf{Zoom in for best view.}
}
\label{abla_space}
\end{figure}

\textbf{Effect of the latent space $\mathcal{G}$. } With verified excellent disentanglement capability of $\mathcal{G}$, we perform ablation experiments to study how it affects the SR performance in PSNR and LPIPS~\cite{zhang2018unreasonable}. As shown in Table~\ref{ablation}, we replace the intermediate latent space $\mathcal{W}$ in the \textit{Base} model with $\mathcal{W+}$, $\mathcal{S}$ and our $\mathcal{G}$ as denoted by \textit{Base+$\mathcal{W+}$}, \textit{Base+$\mathcal{S}$} and \textit{Base+GDM (HMRR)}, respectively. It can be observed that \textit{Base+GDM (HMRR)} achieves the best PSNR and LPIPS, which shows that GDM improves SR by creating a better disentangled latent space to better utilize priors in pre-trained StyleGAN. The experimental results are well aligned with that of qualitative experiments in Fig.~\ref{abla_space} where the model with the proposed $\mathcal{G}$ produces superior face attributes around eyes and teeth.

\begin{figure}[t]
\begin{center}
\includegraphics[width=1\linewidth]{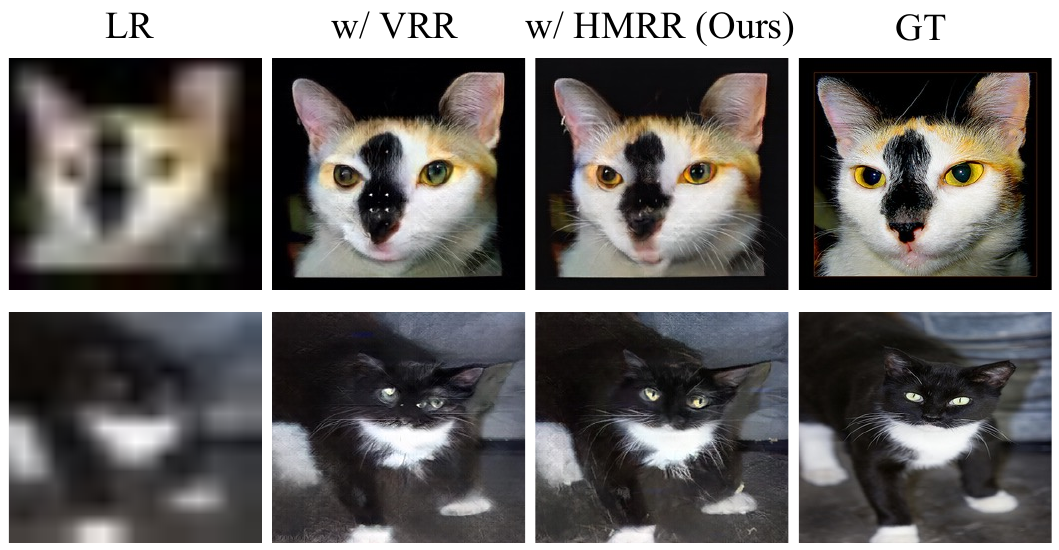}
\end{center}
\caption{
Qualitative ablation studies of our designed hierarchical multi-relation reasoning (HMRR) on 16 $\times$ SR. It can be observed that HMRR helps generate more faithful visual attributes (e.g., eyes) than vanilla relation reasoning (VRR). \textbf{Zoom in for best view.}
}
\label{abla_hmrr}
\end{figure}

\textbf{Effect of hierarchical multi-relation reasoning. } 
To examine the effectiveness of the proposed hierarchical multi-relation reasoning (HMRR), we further train a variant model \textit{Base+GDM (VRR)} which replaces HMRR with the vanilla relation reasoning (VRR). Fig.~\ref{abla_hmrr} shows visual comparisons. It can be observed that the model with the proposed HMRR can yield clearly more faithful image attributes in large-factor SR. The superior SR performance of HMRR is largely attributed to the multi-level attribute relations it captures which enables it produce better attribute disentanglement in latent space. As a comparison, the vanilla relation reasoning in \textit{Base+GDM (VRR)} does not capture the multi-level attribute relations, which leads to degraded disentanglement and large-factor image SR.

\subsection{Ablation Studies on Graph-based Code Generation Module}

\begin{figure}[t]
\begin{center}
\includegraphics[width=1\columnwidth]{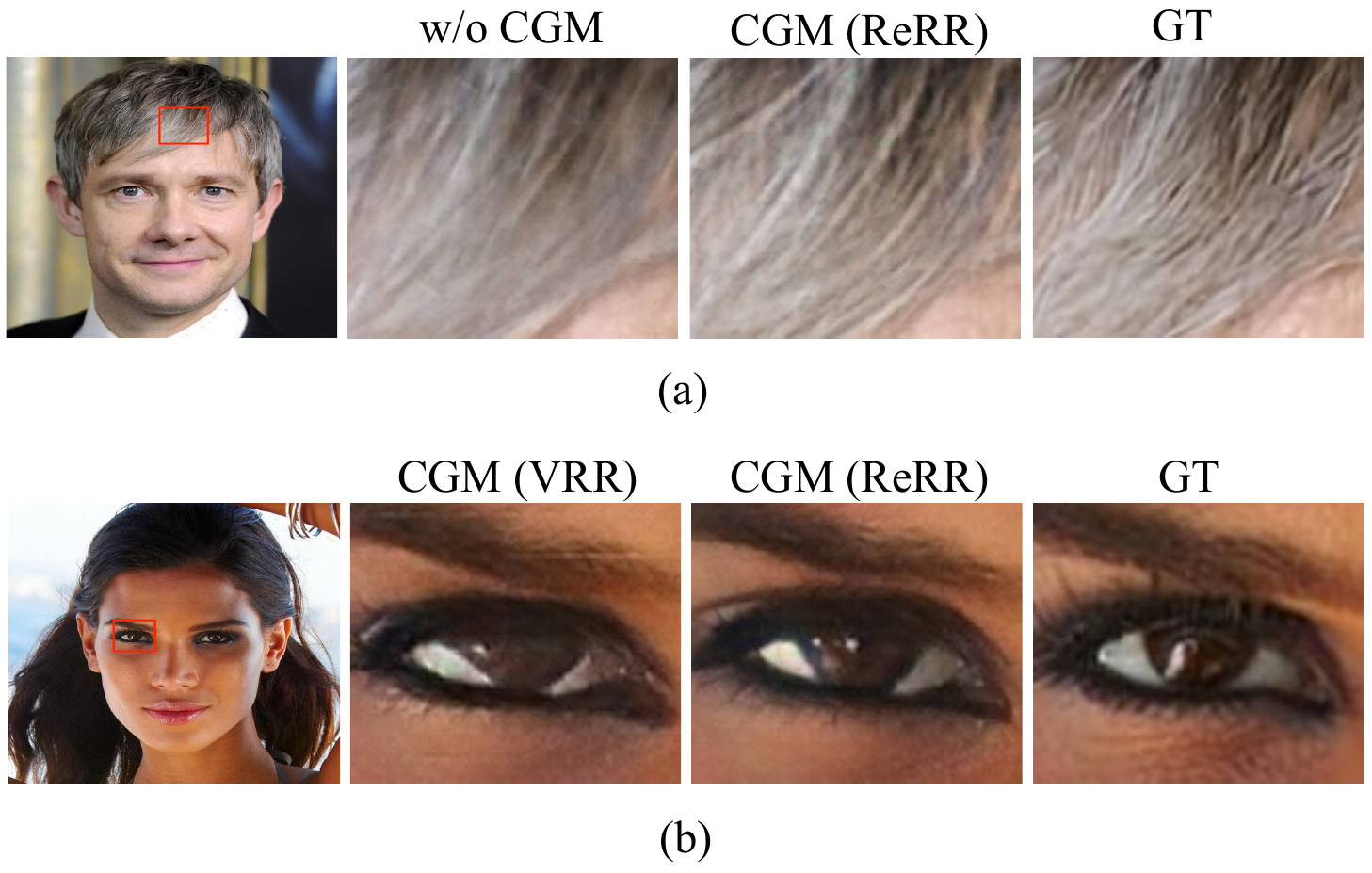}
\end{center}
\caption{Qualitative ablation studies of the proposed CGM and recursive relation reasoning (ReRR) on 16 $\times$ SR. In (a), including CGM can produce more desirable details (e.g., hair) than the model without CGM. In (b), replacing VRR with the proposed ReRR in CGM can further improve detail reconstruction. \textbf{Zoom in for best view.}}
\label{abla_cgm}
\end{figure}

\textbf{Effect of image-specific codes. } We also conduct experiments to examine the contribution of our proposed graph-based code generation module (CGM). As shown in Table~\ref{ablation}, we introduce CGM with the proposed recursive relation reasoning (ReRR) into the model \textit{Base+GDM (HMRR)}, which produces the model \textit{Base+GDM (HMRR) + CGM (ReRR)} that generates image-specific codes to replace stochastic noises. The superior PSNR and LPIPS of the model demonstrate the effectiveness of the image-specific codes provided by CGM. The effectiveness of the proposed CGM can be observed in Fig.~\ref{abla_cgm} (a) as well where the model with CGM can produce more faithful details.

\textbf{Effect of recursive relation reasoning. } We further examine the superiority of our proposed recursive relation reasoning (ReRR) with the comparison between the model \textit{Base+GDM (HMRR)+CGM (ReRR)} and the model \textit{Base+GDM (HMRR)+CGM (VRR)} which replaces the vanilla relation reasoning (VRR) with ReRR. As Table~\ref{ablation} shows, adopting ReRR improves both PSNR and LPIPS consistently. The better performance of ReRR is largely because ReRR considers the relations in previous layers while reasoning relations within the current layer, which improves the reconstruction of fine details. The quantitative results are aligned with the qualitative experiments in Fig.~\ref{abla_cgm} (b). 

\section{Conclusion}
\label{conclusion}

This paper presents LAREN, a GAN-Prior based SR method that aims for high-quality SR reconstruction with large upscale factor via multi-relation reasoning in the latent space. Specifically, a graph disentangled module with hierarchical multi-relation reasoning is designed to construct a superior disentangled latent space for pre-trained GAN to handle visual attribute mismatch and further improve the quality of reconstruction. In addition, we design a graph-based code generation module that progressively yields image-specific codes via recursive relation reasoning to replace stochastic noises (as the input of the pre-trained GAN) and consequently facilitates the generation of faithful image details. Extensive quantitative and qualitative experiments over multiple benchmarks show that the proposed LAREN can achieve high-quality large-factor SR. 

\bibliographystyle{IEEEtran}
\bibliography{cite}

\begin{IEEEbiographynophoto}{Jiahui Zhang}
obtained the B.E. degree in Information Science and Engineering at Shandong University. He is currently pursuing the Ph.D. degree at School of Computer Science and Engineering, Nanyang Technological University. His research interests include computer vision and machine learning, specifically for super-resolution and neural radiance field.
\end{IEEEbiographynophoto}

\begin{IEEEbiographynophoto} {Fangneng Zhan} is a postdoctoral researcher at Max Planck Institute for Informatics. He received the B.E. degree in Communication Engineering and Ph.D. degree in Computer Science \& Engineering from University of Electronic Science and Technology of China and Nanyang Technological University, respectively. His research interests include deep generative models and neural rendering. 
He contributed to the research field by publishing more than 10 articles in prestigious conferences. 
He serves as a reviewer or program committee member for top journals and conferences including TPAMI, TIP, ICLR, NeurIPS, CVPR, ICCV, and ECCV.
\end{IEEEbiographynophoto}

\begin{IEEEbiographynophoto} {Yingchen Yu}
obtained the B.E. degree in Electrical \& Electronic Engineering at Nanyang Technological University, and M.S. degree in Computer Science at National University of Singapore. He is currently pursuing the Ph.D. degree at School of Computer Science and Engineering, Nanyang Technological University under Alibaba Talent Programme. His research interests include computer vision and machine learning, specifically for image synthesis and manipulation.
\end{IEEEbiographynophoto}

\begin{IEEEbiographynophoto} {Rongliang Wu}
received the B.E. degree in Information Engineering from South China University of Technology, and M.S. degree in Electrical and Computer Engineering from National University of Singapore. He is currently pursuing the Ph.D. degree at School of Computer Science and Engineering, Nanyang Technological University. His research interests include computer vision and deep learning, specifically for facial expression analysis and generation.
\end{IEEEbiographynophoto}

\begin{IEEEbiographynophoto} {Xiaoqin Zhang} is a senior member of the IEEE. He received the B.Sc. degree in electronic information science and technology from Central South University, China, in 2005, and the Ph.D. degree in pattern recognition and intelligent system from the National Laboratory of Pattern Recognition, Institute of Automation, Chinese Academy of Sciences, China, in 2010. He is currently a Professor with Wenzhou University, China. He has published more than 100 papers in international and national journals, and international conferences, including IEEE T-PAMI, IJCV, IEEE T-IP, IEEE TNNLS, IEEE T-C, ICCV, CVPR, NIPS, IJCAI, AAAI, and among others. His research interests include in pattern recognition, computer vision, and machine learning.
\end{IEEEbiographynophoto}

\begin{IEEEbiographynophoto} {Shijian Lu} is an Associate Professor in the School of Computer Science and Engineering, Nanyang Technological University. He received his PhD in Electrical and Computer Engineering from the National University of Singapore. His research interests include computer vision and deep learning. He has published more than 100 internationally refereed journal and conference papers. Dr Lu is currently an Associate Editor for the journals of Pattern Recognition and Neurocomputing.
\end{IEEEbiographynophoto}

\vfill

\end{document}